\pdfoutput=1
\documentclass[11pt]{article}

\usepackage{EMNLP2023}

\usepackage{times}
\usepackage{latexsym}

\usepackage[T1]{fontenc}
\usepackage[utf8]{inputenc}

\usepackage{microtype}

\usepackage{inconsolata}

\usepackage{pythonhighlight}
\usepackage{amsmath}
\usepackage{amssymb}

\usepackage{graphicx}
\usepackage{enumitem}
\usepackage{booktabs}

\title{Using Captum to Explain Generative Language Models}

\author{Vivek Miglani*, Aobo Yang*, Aram H. Markosyan, Diego Garcia-Olano, Narine Kokhlikyan \\ \\ Meta AI \\ \{vivekm, aoboyang, amarkos, diegoolano, narine\}@meta.com }

\begin{document}
\maketitle
\def\thefootnote{*}\footnotetext{Denotes equal contribution}\def\thefootnote{\arabic{footnote}}

\begin{abstract}

Captum is a comprehensive library for model explainability in PyTorch, offering a range of methods from the interpretability literature to enhance users' understanding of PyTorch models. In this paper, we introduce new features in Captum\footnote{\url{https://captum.ai}} that are specifically designed to analyze the behavior of generative language models. We provide an overview of the available functionalities and example applications of their potential for understanding learned associations within generative language models.

\end{abstract}

\section{Introduction}

Model interpretability and explainability have become significantly more important as machine learning models are used in critical domains such as healthcare and law. It is insufficient to simply make a prediction through a black-box model and important to better understand why the model made a particular decision. 

Interest in Large Language Models (LLMs) has also grown exponentially in the past few years with the release of increasingly large and more powerful models such as GPT-4~\cite{openai2023gpt4}. A lack of explainability continues to exist despite larger models, and with the use of these models expanding to more and more use-cases, it is increasingly important to have access to tooling providing model explanations.

Captum is an open-source model explainability library for PyTorch providing a variety of generic interpretability methods proposed in recent literature such as Integrated Gradients, LIME, DeepLift, TracIn, TCAV and more \cite{kokhlikyan2020captum}. 

In this work, we discuss newly open-sourced functionalities in Captum v0.7 to easily apply explainability methods to large generative language models, such as GPT-3.

\section{Attribution Methods}

One important class of explainability methods is attribution or feature importance methods, which output a score corresponding to each input feature's contribution or impact to a model's final output.

Formally, given a function $f: \mathbb{R}^d \rightarrow \mathbb{R}$, where $f\in \mathbb{F}$ and $X\in \mathbb{R}^d$ is a single input vector consisting of $d$ dimensions or features, an attribution method is defined as a function
$\phi : \mathbb{F} \times \mathbb{R}^d \rightarrow \mathbb{R}^d.$ Each element in the attribution output corresponds to a score of the contribution of corresponding feature $i\in D$, where $D$ denotes the set of all feature indices $D = \{1,2,...,d\}$.

Many attribution methods also require a baseline or reference input $B \in \mathbb{R}^d$ defining a comparison input point to measure feature importance with respect to. 

We utilize the notation $X_S$ to denote selecting the feature values with indices from the set $S \subseteq D $ and the remaining indices from $B$. Formally, the value of feature $i$ in $X_S$ is $(X_S)_i =  I_{i \in S} X_i + I_{i \notin S} B_i$, where $I$ is the indicator function. 

In this section, we provide background context on attribution methods available in Captum. These methods can be categorized broadly into (i) perturbation-based methods, which utilize repeated evaluations of a black-box model on perturbed inputs to estimate attribution scores, and (ii) gradient-based methods, which utilize backpropagated gradient information to estimate attribution scores. Perturbation-based methods do not require access to model weights, while gradient-based models do.   

\subsection{Perturbation-Based Methods}

\subsubsection{Feature Ablation}

The most straightforward attribution is feature ablation, where each feature is substituted with the corresponding element of the baseline feature vector to estimate the corresponding importance.

Formally, this method is defined as
\begin{equation}
    \phi_i (f, X) = f(X) - f(X_{D\setminus \{i\}})
\end{equation}

Feature Ablation has clear advantages as a simple and straightforward method, but the resulting attributions may not fully capture the impacts of feature interactions since features are ablated individually.

\subsubsection{Shapley Value Sampling}

Shapley Values originated from cooperative game theory as an approach to distribute payouts fairly among players in a cooperative game.  Analogously, in the attribution setting, Shapley Values assign scores to input features, with payouts corresponding to a feature's contribution to the model output. Shapley Values satisfy a variety of theoretical properties including efficiency, symmetry and linearity.
Formally, this method is defined as

\begin{equation}
\begin{split}
    \phi_i (f, X) = \sum_{S \subseteq D \setminus \{i\}} \left[ \frac{|S|!(|D| - |S| - 1)!}{|D|!}  \right. \\
    \left. f(X_{S \cup \{i\}}) - f(X_{S}) \right]
\end{split}
\end{equation}

While computing this quantity exactly requires an exponential number of evaluations in the number of features, we can estimate this quantity using a sampling approach \cite{castro2009polynomial}. The approach involves selecting a permutation of the $d$ features and adding the features one-by-one to the original baseline. The output change as a result of adding each feature accounts for its contribution, and averaging this over sampled perturbations results in an unbiased estimate of Shapley Values.

\subsubsection{LIME}

LIME or Locally Interpretable Model Explanations proposes a generic approach to sample points in the neighborhood of the input point $X$ and train an interpretable model (such as a linear model) based on the results of the local evaluations \cite{lime}.  

This method proposes reparametrizing the input space to construct interpretable features such as super-pixels in images and then evaluating the original model on a variety of perturbations of the interpretable features. The method can be utilized with any perturbation sampling and weighting approaches and interpretable model / regularization parameters. The interpretable model can then be used as an explanation of the model's behavior in the local region surrounding the target input point. For a linear model, the coefficients of this model can be considered as attribution scores for the corresponding feature.

\subsubsection{Kernel SHAP}
Kernel SHAP is a special case of the LIME framework, which sets the sampling approach, intepretable model, and regularization in a specific way such that the results theoretically approximate Shapley Values \cite{kernelshap}.

\subsection{Gradient Based Methods}

\subsubsection{Saliency}
Saliency is a simple gradient-based approach, utilizing the model's gradient at the input point as the corresponding feature attribution \cite{simonyan2013deep}. This method can be understood as taking a first order approximation of the function, in which the gradients would serve as the coefficients of each feature in the model.
\begin{equation}
\phi_i (f, X) = f'(X)    
\end{equation}

\subsubsection{Integrated Gradients}
Integrated Gradients estimates attribution by computing the path integral of model gradients between the baseline point and input point \cite{sundararajan2017axiomatic}. This approach has been shown to satisfy desirable theoretical properties including sensitivity and implementation invariance. Formally, the method can be defined as
\begin{equation}
\begin{split}
\phi_i (f, X) = (X_i - B_i) \\ \times \int_{\alpha = 0}^1 \frac{f'(B + (X - B)\alpha )}{\partial x_i} d \alpha
\end{split}
\end{equation}

\subsubsection{Other Gradient-Based Methods}
Other popular gradient-based attribution methods include DeepLift and Layerwise Relevance Propogation (LRP) \cite{ shrikumar2017learning, bach2015pixel}. These methods both require a backward pass of the model on the original inputs but customize the backward propagation of specific functions, instead of using their default gradient functions.

\begin{figure*}[h]
\includegraphics[width=0.99\linewidth]{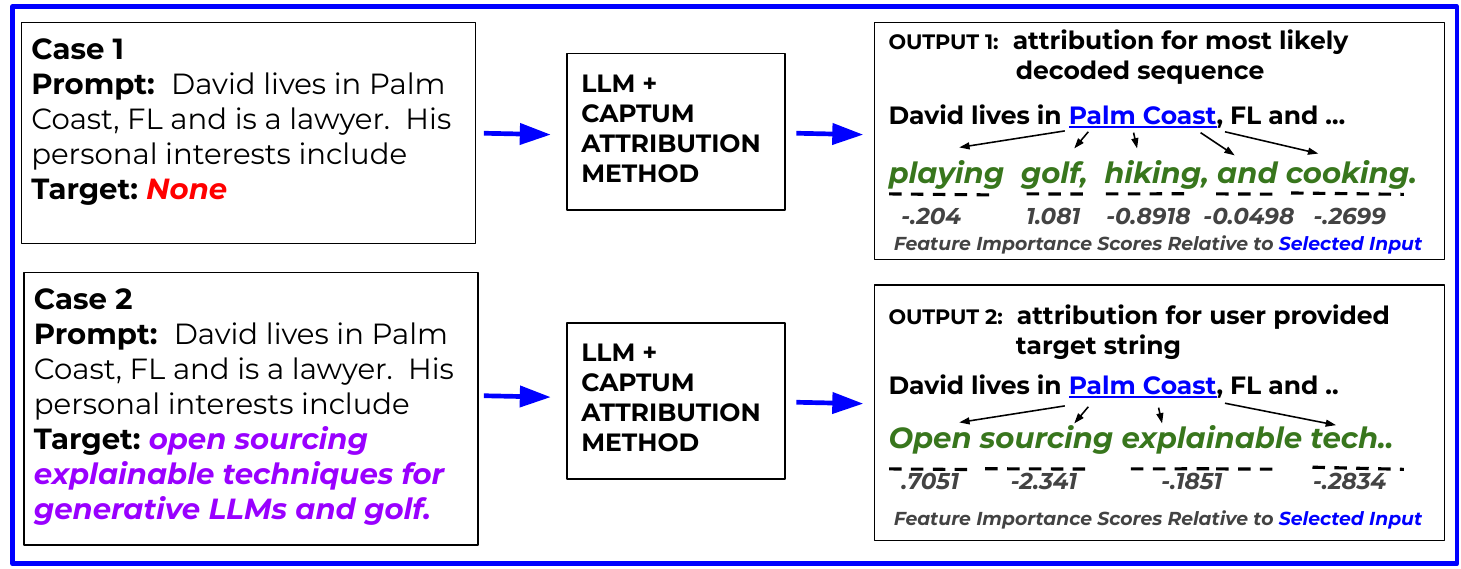}
  \caption{Example of applying Captum attribution methods to analyze the input prompt's effect on the output of LLMs, showing two types of target strings accepted by Captum attribution API and token level attribution outputs for both with respect to the input "Palm Coast". In Case 1, no Target string is provided, so attributions are provided for the most likely decoded sequence. In Case 2, attributions are provided for the chosen target output.}
  \label{viz_target_selection}
\end{figure*}

\section{Language Model Attribution}
In Captum v0.7, we propose new functionalities to apply the attribution methods within Captum to analyze the behaviors of LLMs. Users can choose any interested tokens or segments from the input prompt as features, e.g., "Palm Coast" in the example shown in Figure \ref{viz_target_selection}, and use attribution methods to quantify their impacts to the generation targets, which can be either a specified output sequence or a likely generation from the model.

\subsection{Perturbation-Based Methods}
We introduce simple APIs to experiment with perturbation-based attribution methods including Feature Ablation, Lime, Kernel SHAP and Shapley Value Sampling.

We prioritize ease-of-use and flexibility, allowing users to customize the chosen features for attribution, mask features into groups as necessary, and define appropriate baselines to ensure perturbed inputs remain within the natural data distribution.
\begin{figure*}[hbt!]
\begin{python}
from captum.attr import FeatureAblation, LLMAttribution, TextTemplateFeature

fa = FeatureAblation(model)
llm_attr = LLMAttribution(fa, tokenizer)

inp = TextTemplateFeature(
    # the text template
    "{} lives in {}, {} and is a {}. {} personal interests include", 
    # the values of the features
    ["Dave", "Palm Coast", "FL", "lawyer", "His"],
    # the reference baseline values of the features
    baselines=["Sarah", "Seattle", "WA", "doctor", "Her"],
)
llm_attr.attribute(inp)
\end{python}
\caption{Example of applying Captum with a list of features in a text template}
\label{perturbation-list}
\end{figure*}

\begin{figure*}
\begin{python}
inp = TextTemplateFeature(
    "{name} lives in {city}, {state} and is a {occupation}. {pronoun} personal interests include", 
   {"name":"Dave", "city": "Palm Coast", "state": "FL", "occupation":"lawyer", "pronoun":"His"}, 
   baselines={"name":"Sarah", "city": "Seattle", "state": "WA", "occupation":"doctor", "pronoun":"Her"}
)
attr_result = llm_attr.attribute(inp, target="playing golf, hiking, and cooking.")
attr_result.plot_token_attr()
\end{python}
\caption{Example of applying Captum with a dictionary of features in a text template and a specific target, and visualize the token attribution}
\label{perturbation-dict}
\end{figure*}

In Figure \ref{perturbation-list}, we demonstrate an example usage of the LLMAttribution API for the simple prompt \emph{"Dave lives in Palm Coast, FL and is a lawyer. His personal interests include"}. Providing this input prompt to a language model to generate the most likely subsequent tokens, we can apply Captum to understand the impact of different parts of the prompt string on the model generation. Figure \ref{perturbation-dict} presents a more customized usage where we use the same function to understand a specific output we are interested in (\emph{"playing golf, hiking, and cooking."}).

\subsubsection{Defining Features}
Users are able to define and customize 'features' for attribution in the prompt text. The simplest approach would be defining the features as individual tokens.

Unfortunately, in many cases, it doesn't make sense to perturb individual tokens, since this may no longer form a valid sentence in the natural distribution of potential input sequences. For example, perturbing the token "Palm" in the above sentence would result in a city name that is not in the natural distribution of potential cities in Florida, which may lead to unexpected impacts on the perturbed model output. Moreover, tokenizers used in modern LLMs may further break a single word in many cases. For example, the tokenizer can break the word \emph{"spending"} into \emph{"\_sp"} and \emph{"ending"}.

The API provides flexibility to define units of attribution as custom interpretable features which could be individual words, tokens, phrases, or even full sentences. For example, in Figure~\ref{perturbation-list}, we select the relevant features to be the name, city, state, occupation, and pronoun in the sentence prompt and desire to determine the relative contribution of these contextual features on the model's predicted sentence completion.

Users can define the units for attribution as a list or dictionary of features and provide a format string or function to define a mapping from the attribution units to the full input prompt as shown in Figure~\ref{perturbation-dict}.

\subsubsection{Baselines}
The baseline choice is particularly important for computing attribution for text features, as it serves as the reference value used when perturbing the chosen feature. The perturbation-based feature API allows defining custom baselines corresponding to each input feature.

It is recommended to select a baseline which fits the context of the original text and remains within the natural data distribution. For example, replacing the name of a city with another city ensures the sentence remains naturally coherent, but allows measuring the contribution of the particular city selected.

In addition to a single baseline, the Captum API also supports providing a distribution of baselines, provided as either a list or function to sample a replacement option. For example, in the example above, the name "Dave" could be replaced with a sample from the distribution of common first names to measure any impact of the name "Dave" in comparison to the chosen random distribution as shown in Figure~\ref{assoc-example}.

\subsubsection{Masking Features}

Similar to the underlying Captum attribution methods, we support feature masking, which enables grouping features together to perturb as a single unit and obtain a combined, single attribution score. This functionality may be utilized for highly correlated features in the text input, where it often makes sense to ablate these features together, rather than independently.

For example, in Figure~\ref{perturbation-list}, the feature pairs (city, state) and (name, pronoun) are often highly correlated, and thus it may make sense to group them and consider them as a single feature.

\subsubsection{Target Selection}
For any attribution method, it is also necessary to select the target function output for which attribution outputs are computed. Since language models typically output a probability distribution over the space of tokens for each subsequently generated token, there are numerous choices for the appropriate target.

By default, when no target is provided, the target function behavior is for the attribution method to return attributions with respect to the most likely decoded token sequence.

When a target string is provided, the target function is the log probability of the output sequence from the language model, given the input prompt. For a sequence with multiple tokens, this is numerically computed through the sum of the log probabilities of each token in the target string.  Figure~\ref{viz_target_selection} shows these two input use cases and shows token level attribution relative to an input subsequence for both.

We also support providing a custom function on the output logit distribution, which allows attribution with respect to an alternate quantity such as the entropy of the output token distribution.

\begin{figure*}
\begin{python}
from captum.attr import LayerIntegratedGradients, TextTokenFeature

ig = LayerIntegratedGradients(model, "model.embed_tokens")
llm_attr = LLMGradientAttribution(ig, tokenizer)

inp = TextTokenFeature("Dave lives in Palm Coast, FL and is a lawyer. His personal interests include", tokenizer)
llm_attr.attribute(inp)
\end{python}
\caption{Example of applying Captum with a gradient-based approach}
\label{gradient-attr}
\end{figure*}

\subsection{Gradient-Based Methods}
Captum 0.7 also provides a simple API to apply gradient-based methods to LLMs. Applying these methods to language models is typically more challenging than for models with dense feature inputs, since embedding lookups in LLMs are typically non-differentiable functions, and gradient-based attributions need to be obtained with respect to embedding outputs.
Captum allows these attributions to be aggregated across embedding dimensions to obtain per-token attribution scores. Figure \ref{gradient-attr} demonstrates an example of applying Integrated Gradients on a sample input prompt.

\subsection{Visualization}

\begin{figure}[h]
\includegraphics[width=0.95\linewidth]{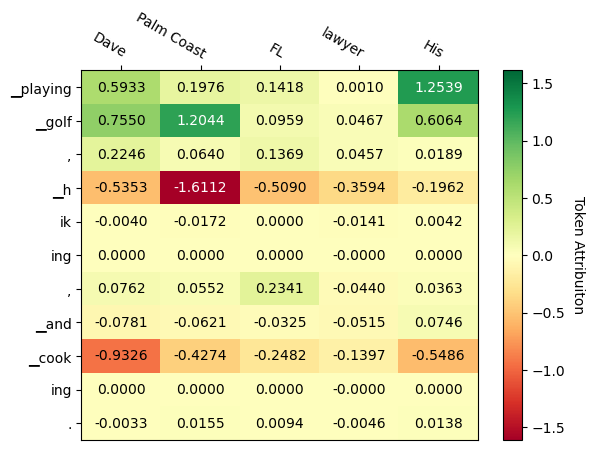}
  \caption{Text Attribution Visualization Example}
  \label{viz_example_figure}
\end{figure}

We also open source utilities for easily visualizing the attribution outputs from language models. Figure~\ref{perturbation-dict} shows how to use the utilities to visualize the attribution result, and Figure~\ref{viz_example_figure} demonstrates the heatmap plotted with the prompt along the top, the target string along the left side and feature importance scores in each cell.

% \begin{figure*}
% \begin{python}
% attr_viz = LLMAttrViz(**attr_result)
% attr_viz.plot_seq_attr()
% attr_viz.plot_token_attr()
% \end{python}
% \caption{Visualizing Captum Sequence Attributions}
% \label{viz-example}
% \end{figure*}

\begin{table}[!h]
    \begin{center}
        \resizebox{\columnwidth}{!}{\begin{tabular}{ l l l l}
    \toprule
        (Feature) Value & Golfing & Hiking & Cooking \\ 
        \midrule
        (Name) Dave & 0.4660 & -0.2640 & -0.4515 \\
        (City) Palm Coast & 1.0810 & -0.8762 & -0.2699 \\  
        (State) FL & 0.6070 & -0.3620 & -0.3513 \\
        (Occupation) lawyer & 0.7584 & -0.1966 & 0.0331 \\
        (Pronoun) His & 0.2217 & -0.0650 & -0.2577 \\
        \bottomrule
        \end{tabular}}
    \end{center}
    \caption{Associations between input and generated text features}
    \label{tab:association}
\end{table}

\section{Applications}

In this section, we discuss two applications of the attribution methods described above in different contexts.  These applications provide additional transparency as well as contribute to a better understanding of a model's learned associations and robustness.

\subsection{Understanding Model Associations}

This perturbation-based tooling can be particularly useful for improved understanding of learned associations in LLMs.

Consider the example prompt:
\begin{itemize}[leftmargin=*]
    \item[]  \textit{``David lives in Palm Coast, FL and is a lawyer. His personal interests include ''}
\end{itemize}

We can define features including gender, city, state and occupation. Obtaining attributions on these features against the subsequent target
\begin{itemize}[leftmargin=*]
    \item[]  \textit{``playing golf, hiking, and cooking. ''}
\end{itemize}
allows us to better understand why the model predicted these personal interests and how each feature correlates with each of these interests. The model might be associating this activity as a common hobby for residents in the specific city or as an activity common to lawyers. Through this choice of features, we can obtain a better understanding of why the model predicted these particular hobbies and how this associates with the context provided in the prompt. 

We apply Shapley Value Sampling to better understand how each of the features contributed to the prediction. The corresponding code snippet is shown in the Appendix in Figure 6. Table \ref{tab:association} presents the effects of each feature on the LLM's probability of outputting the corresponding interest, with positive and negative values indicating increases and decreases of the probability respectively. We can therefore identify some interesting and even potentially biased associations. For example, the male name and pronoun, i.e., "Dave" and "His", have positive attribution to "golfing" but negative attribution to "cooking".

\subsection{Evaluating Effectiveness of Few-Shot Prompts}

Significant prior literature has demonstrated the ability of LLMs to serve as few-shot learners \cite{brown2020language}. We utilize Captum's attribution functionality to better understand the impact and contributions of few-shot examples to model predictions. Table~\ref{tab:example_prompts} demonstrates four example few shot prompts and corresponding attribution scores when predicting  positive sentiment for "I really liked the movie, it had a captivating plot!" movie review.

%
%\begin{itemize}[leftmargin=*]
%    \item[]  \textit{``Here are some examples of movie reviews and classification of whether they were Positive or Negative:}
        
%        \textit{`The movie was ok, the actors weren't great' -> Negative} \\
%        \textit{`I loved it, it was an amazing story!' -> Positive} \\
%        \textit{`Total waste of time!!' ->  Negative } \\
%        \textit{`Won't recommend' ->  Negative}
        
%        \textit{Decide if the following movie review enclosed in quotes is Positive or Negative:}
        
%        \textit{`I really liked the movie, it had a captivating plot!'}
        
%        \textit{Only output Positive or Negative.''}
%\end{itemize}

% \begin{center}    
% \fbox{\begin{minipage}{15em}
% Here are some examples of movie reviews and classification of whether they were Positive or Negative.

% 'The movie was ok, the actors weren't great' -> Negative 

% 'I loved it, it was an amazing story!' -> Positive 

% 'Total waste of time!!' ->  Negative 

% 'Won't recommend' ->  Negative

% Decide if the following movie review enclosed in quotes is Positive or Negative. Output only either Positive or Negative. 

% 'I really liked the movie, it had a captivating plot!'
% \end{minipage}}
% \end{center}

Here we aim to understand the impact of each example prompt on the \textit{Positive} sentiment prediction.
The LLM is asked to predict positive or negative sentiment using the following prompt:

\begin{itemize}[leftmargin=*]
    \item[]  \textit{``Decide if the following movie review enclosed in quotes is Positive or Negative. Output only either Positive or Negative:}

    \textit{`I really liked the movie, it had a captivating plot!' ''}
\end{itemize}     

% \begin{center}    
% \fbox{\begin{minipage}{15em}
% Decide if the following movie review enclosed in quotes is Positive or Negative. Output only either Positive or Negative. 

% 'I really liked the movie, it had a captivating plot!'
% \end{minipage}}
% \end{center}

We consider each of the provided example prompts as features and we utilize zero-shot prompt as a baseline in the attribution algorithm. The detailed implementation can be found in Appendix in Figure 7.

We obtain results as shown in Table \ref{tab:example_prompts} by applying Shapley Values. Surprisingly, the results suggest that all the provided examples actually reduced confidence in the prediction of "Positive".

\begin{table}[!h]
    \begin{center}
    \begin{tabular}{ p{2.25in}p{0.5in}}
        \toprule
        Example & Shapley Value \\ 
        \midrule
        'The movie was ok, the actors weren't great' -> Negative  & -0.0413  \\
        'I loved it, it was an amazing story!' -> Positive & -0.2751 \\ 
        'Total waste of time!!' ->  Negative  & -0.2085 \\ 
        'Won't recommend' ->  Negative  & -0.0399 \\
        \bottomrule
    \end{tabular}
    \end{center}
    \caption{Example prompts' contribution to model response "Positive."}
    \label{tab:example_prompts}
\end{table}

\section{Related Work}
Numerous prior works have developed and investigated attribution methods with a variety of properties, but few efforts have been made to develop open-source interpretability tools providing a variety of available methods, particularly for the text domain. Captum was initially developed to fill this gap and provide a centralized resource for recent interpretability methods proposed in literature \cite{kokhlikyan2020captum}.

Ecco and inseq are two libraries that have provided attribution functionalities for text / language models \cite{inseq, alammar-2021-ecco}, and both libraries are built on top of the attribution methods available in Captum. These libraries primarily focus on gradient-based attribution methods, which provide token-level attribution based on gradient information.

In contrast, our main contribution in this work has been a focus on perturbation-based methods and providing flexibility on aspects such as feature definition, baseline choice and masking. We do not necessarily expect that these attribution methods provide a score for each token individually, which is typically the case for gradient-based methods. This shift in structure allows us to generalize to a variety of cases and allows the users to define features for attribution as it fits best.

Some prior work on attribution methods have also demonstrated limitations and counterexamples of these methods in explaining a model's reliance on input features, particularly with gradient-based attribution methods \cite{sanity}. 

Perturbation-based methods generally have an advantage of being justifiable through the model's output on counterfactual perturbed inputs. But perturbing features by removing individual tokens or replacing them with pad or other baseline tokens may result in inputs outside of the natural data distribution, and thus, model outputs in this region may not be accurate. The tools developed have been designed to make it easier for developers to select features, baselines, and masks which can ensure perturbations remain within the natural data distribution in order to obtain more reliable feature attribution results.

Recent advances in data augmentation \cite{pluščec2023data} for natural language processing have led to the development of a number of open-source libraries~\cite{wang-etal-2021-textflint, papakipos2022augly, Zeng_2021, morris2020textattack, ma2019nlpaug, dhole2022nlaugmenter, wu2021polyjuice}. Among many functionalities, these libraries provide a rich set of text perturbations. Some libraries have specific focus, e.g. perturbing demographic references \cite{qian-etal-2022-perturbation}. An interesting direction of future work will be the extension of our present API to provide fully automated feature and baseline selections, allowing users to simply provide an input string and automatically identify appropriate text features and corresponding baselines for attribution.   

\section{Conclusion}
In this work, we introduce new features in the open source library Captum that are specifically designed to analyze the behavior of generative LLMs. We provide an overview of the available functionalities and example applications of their potential in understanding learned associations and evaluating effectiveness of few-shot prompts within generative LLMs. We demonstrate examples for using perturbation and gradient-based attribution methods with Captum which highlight the API's flexibility on aspects such as feature definition, baseline choice and masking. This flexibility in structure allows users to generalize to a variety of cases, simplifying their ability to conduct explainability experiments on generative LLMs.

In the future, we plan to expand our API for additional automation in baseline and feature selection as well as incorporate other categories of interpretability techniques for language models. Runtime performance optimization of attribution algorithms is another area of research that could be beneficial for the OSS community.

\bibliography{custom}
\bibliographystyle{acl_natbib}

\appendix

\section{Appendix}

\begin{figure*}
\begin{python}
from captum.attr import ShapleyValueSampling, LLMAttribution, TextTemplateFeature, ProductBaselines

svs = ShapleyValueSampling(model) 
baselines = ProductBaselines(
    {
        ("name", "pronoun"): [("Sarah", "Her"), ("John", "His")],
        "city": ["Seattle", "Boston"], 
        "state": ["WA", "MA"], 
        "occupation": ["doctor", "engineer", "teacher", "technician", "plumber"], 
    }
)

llm_attr = LLMAttribution(svs, tokenizer)

inp = TextTemplateFeature(
    "{name} lives in {city}, {state} and is a {occupation}. {pronoun} personal interests include", 
    {"name":"Dave", "city": "Palm Beach", "state": "FL", "occupation":"lawyer", "pronoun":"His"},
    baselines=baselines,
)

attr_result = llm_attr.attribute(inp, target="playing golf, hiking, and cooking.")
\end{python}
  \caption{Applying Captum for the model associations example}
  \label{assoc-example}
\end{figure*}

\begin{figure*}
\label{few_shot}
\begin{python}
from captum.attr import ShapleyValues, LLMAttribution, TextTemplateFeature

sv = ShapleyValues(model) 
llm_attr = LLMAttribution(sv, tokenizer)

def prompt_fn(*examples):
    main_prompt = "Decide if the following movie review enclosed in quotes is Positive or Negative:\n'I really liked the Avengers, it had a captivating plot!'\nReply only Positive or Negative."
    subset = [elem for elem in examples if elem]
    if not subset:
        prompt = main_prompt
    else:
        prefix = "Here are some examples of movie reviews and classification of whether they were Positive or Negative:\n"
        prompt = prefix + "\n".join(subset) + "\n" + main_prompt
    return "[INST] " + prompt + "[/INST]"

input_examples = [
    "'The movie was ok, the actors weren't great' -> Negative", 
    "'I loved it, it was an amazing story!' -> Positive",
    "'Total waste of time!!' -> Negative", 
    "'Won't recommend' -> Negative",
]
inp = TextTemplateFeature(prompt_fn, input_examples)

attr_result = llm_attr.attribute(inp)
\end{python}
  \caption{Applying Captum for the few-shot prompt example}
\end{figure*}

\end{document}